\documentclass[runningheads]{llncs}

\usepackage[T1]{fontenc}
\usepackage{graphicx}
\usepackage{bbding}
\usepackage[misc]{ifsym}
\usepackage[ruled,linesnumbered]{algorithm2e}
\usepackage[colorlinks=true, linkcolor=blue, citecolor=blue, urlcolor=blue]{hyperref}
\usepackage[table]{xcolor}
\usepackage{multirow}
\usepackage{mathrsfs}
\usepackage{booktabs}
\usepackage{threeparttable}
\usepackage{amsmath,amssymb}
\usepackage{bbm}
\usepackage{bbding}

\begin{document}
\title{Targeted False Positive Synthesis via Detector-guided Adversarial Diffusion Attacker for Robust Polyp Detection}
%


\author{Quan Zhou$^{\dag}$\inst{1} 
	\and
        Gan Luo$^{\dag}$\inst{1} 
	\and
        Qiang Hu\textsuperscript{\textrm{(\Letter)}} \inst{2} 
	\and
        Qingyong Zhang\inst{1} 
	\and
        Jinhua Zhang\inst{3} 
	\and
	Yinjiao Tian\inst{3} 
	\and
	Qiang Li\inst{2} 
        \and
	Zhiwei Wang\textsuperscript{\textrm{(\Letter)}} \inst{2}} 

\titlerunning{Detector-guided Adversarial Diffusion Attacker}
\authorrunning{Q. Zhou et al.}
%
\institute{Wuhan University of Technology \and WNLO, Huazhong University of Science and Technology \and Changzhou United lmaging Healthcare Surgical Technology Co., Ltd.\\
\email{\{zhouquan910, ganluo\}@whut.edu.cn, \{huqiang77, zwwang\}@hust.edu.cn}}


	%
\maketitle              
\def\thefootnote{$\dag$}\footnotetext{Equal~contribution; \textrm{\Letter}~corresponding author.}
\begin{abstract}
    Polyp detection is crucial for colorectal cancer screening, yet existing models are limited by the scale and diversity of available data. While generative models show promise for data augmentation, current methods mainly focus on enhancing polyp diversity, often overlooking the critical issue of false positives. In this paper, we address this gap by proposing an adversarial diffusion framework to synthesize high-value false positives. The extensive variability of negative backgrounds presents a significant challenge in false positive synthesis. To overcome this, we introduce two key innovations: First, we design a regional noise matching strategy to construct a negative synthesis space using polyp detection datasets. This strategy trains a negative-centric diffusion model by masking polyp regions, ensuring the model focuses exclusively on learning diverse background patterns. Second, we introduce the Detector-guided Adversarial Diffusion Attacker (DADA) module, which perturbs the negative synthesis process to disrupt a pre-trained detector's decision, guiding the negative-centric diffusion model to generate high-value, detector-confusing false positives instead of low-value, ordinary backgrounds. Our approach is the first to apply adversarial diffusion to lesion detection, establishing a new paradigm for targeted false positive synthesis and paving the way for more reliable clinical applications in colorectal cancer screening. Extensive results on public and in-house datasets verify the superiority of our method over the current state-of-the-arts, with our synthesized data improving the detectors by at least $2.6\%$ and $2.7\%$ in F1-score, respectively, over the baselines. Codes are at \href{https://github.com/Huster-Hq/DADA}{https://github.com/Huster-Hq/DADA}.
    
\end{abstract}

\section{Introduction}
Colorectal cancer (CRC) remains one of the most prevalent and deadly malignancies worldwide, presenting significant challenges to global public health \cite{granados2017colorectal}. Early and accurate screening of colonic polyps during colonoscopy is critical in reducing the morbidity and mortality associated with CRC. However, polyp screening is often hindered by the variability in clinicians' skills and experience levels, which underscores the urgent need for automatic polyp detection.

Recent advancements in polyp detection have primarily focused on innovations in network architectures, such as Transformer-based \cite{yoo2024real}, YOLO-based \cite{pacal2021robust}, R-CNN-based \cite{yang2020colon,chen2021self}. Further progress has been made by leveraging temporal relationships to improve detection accuracy, utilizing techniques like 3D models \cite{puyal2022polyp}, optical flow correlations \cite{zheng2019polyp}, and tracking modules \cite{yu2022end}. Despite these advances, polyp detection still faces substantial challenges due to the variability in polyp morphology and the complex, dynamic nature of the colon environment. While certain methods address specific issues, such as reflection artifacts \cite{kayser2020understanding,hu2024sali} and size variability \cite{souaidi2022multi}, the lack of comprehensive and diverse datasets remains a major barrier to further progress \cite{adjei2022examining}.

In recent years, breakthroughs in generative models, particularly denoising diffusion probabilistic models (DDPMs), have provided new opportunities to enhance object detection by generating challenging, high-value training data. For example, DS-GAN \cite{bosquet2023full} has proven effective in generating realistic small targets, addressing the challenges of detecting small objects. In polyp detection, Qadir \emph{et al.} \cite{qadir2022simple} introduced a conditional GAN-based framework for synthesizing synthetic polyp images, improving both detection and segmentation accuracy. Similarly, Adjei \emph{et al.} \cite{adjei2022examining} integrated a modified pix2pix framework with traditional augmentation techniques to boost model performance. More recently, ControlPolypNet \cite{sharma2024controlpolypnet} employed user-controllable inputs to generate clinically relevant polyp images, expanding datasets and improving segmentation outcomes.

However, these methods primarily focus on generating positive samples, with relatively little attention to the issue of \textbf{false positives}. In a 15- to 20-minute polyp screening, negative samples significantly outnumber positive ones, with current systems producing an average of five false positives per minute~\cite{spadaccini2022comparing}. This excessive number of false positives can disrupt clinicians' focus \cite{holzwanger2021benchmarking}, a phenomenon known as the `crying wolf effect'. It not only increases the workload of clinicians but also introduces the risk of unnecessary medical interventions and costly diagnostic errors \cite{mori2021addressing}. Furthermore, individual variability and differences in bowel preparation result in highly complex colon environments, further exacerbating the false positive problem and making reliable detection more difficult.

In this paper, we address a critical gap for the first time by proposing a novel method for targeted false positive generation. Our approach seeks to reduce false positives by generating challenging negative training samples that mimic polyp-like features, which can confuse the detector. Specifically, our method consists of two key components: a background-only DDPM as the base negative generator and a Detector-guided Adversarial Diffusion Attacker (DADA) module.
The base negative generator is trained on \emph{off-the-shelf} polyp detection datasets but employs a regional noise matching strategy to decouple the polyp regions from the DDPM's training.
This strategy ensures that the generator focuses solely on learning diverse non-polyp background visual patterns to avoid leak of polyp information into the subsequent generation of polyp-like interferences. During the denoising process, the DADA module, inspired by adversarial attack principles, establishes a gradient backpropagation pathway to enable the injection of adversarial perturbations into the denoising process, effectively confusing the detector's decision boundary and guiding the generation of general low-value negatives into high-value false positive samples.

In summary, our major contributions are as follows:
\begin{itemize} 
        \item We propose a novel image synthesis method that integrates diffusion models with adversarial attacks, focusing on generating high-value negative samples capable of effectively misleading polyp detectors.
    \item We propose a background-only denoiser learning to generate pure negative patterns from \emph{off-the-shelf} polyp detection datasets, and a Detector-guided Adversarial Diffusion Attacker (DADA) module guiding the denoising to high-value realistic false positive ones by attacking a well-trained detector. 
    \item Extensive experiments demonstrate that our method achieves state-of-the-art performance on both the public Kvasir and our in-house dataset, with improvements of at least $2.6\%$ and $2.7\%$ in F1-score, respectively, over the baselines.
\end{itemize}

\section{Method}
\begin{figure}[t]
    \centering
    \includegraphics[width=\textwidth]{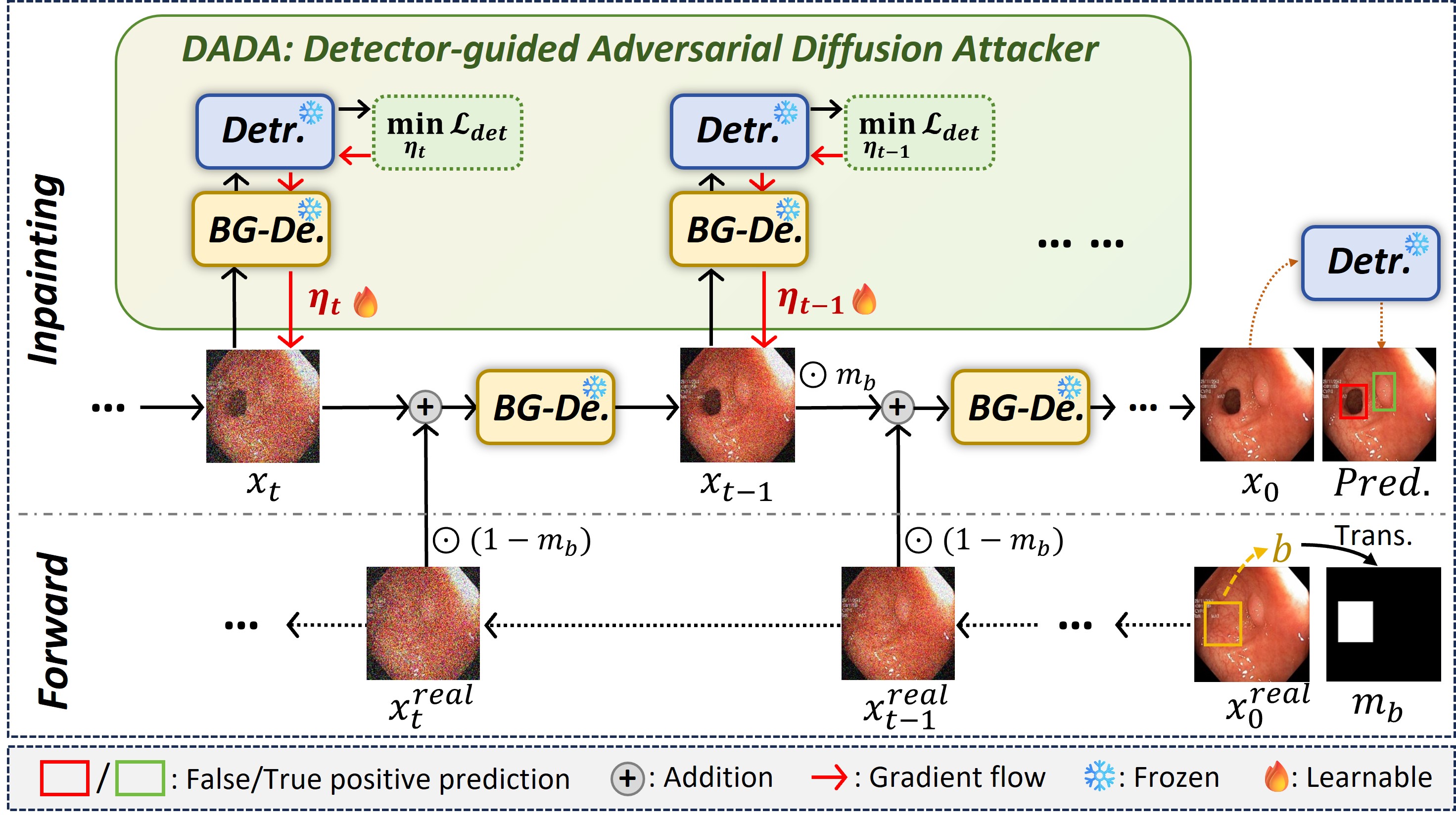}
    \caption{The inference pipeline consists of three key modules: the background-only denoiser (BG-De), a well-trained polyp detector (Detr), and the detector-guided adversarial diffusion attacker (DADA). Inpainting is used to generate new negative samples in a user-specified region of the real image. BG-De samples from the pure \emph{negative} distribution, while DADA guides the sampling toward visual patterns that elicit a \emph{positive} response from the detector. Note that the training of both BG-De and the detector is not shown in this figure.}
\label{fig1}
\end{figure}

Fig.~\ref{fig1} shows the overall of our adversarial diffusion framework, which mainly consists of three key modules: (1) a well-trained polyp detector, (2) a background-only denoiser (BG-De), and (3) a detector-guided adversarial diffusion attacker (DADA). During inference, the iterative process alternates between denoising through BG-De, evaluating their potential for false positive induction via the detector's prediction, and applying DADA-computed adversarial perturbations to confuse the detector in a pre-defined region of interest.

\subsection{Background-only Denoiser as Base Negative Generator using Regional Noise Matching} \label{sec:2.1}
BG-De is a variant of the Diffusion Probabilistic Model (DDPM)~\cite{ho2020denoising}, designed to selectively denoise background regions. During the forward diffusion process, BG-De adds Gaussian noise to an image $x_0$ over $t$ time steps:~$x_t=\sqrt{\overline{\alpha}_t}x_{0}+\sqrt{1-\overline{\alpha}_t}\epsilon_{x_t}$, where $\overline{\alpha}_t = \prod^t_{i=1}\alpha_i$, and $\alpha_{1:T}$ are hyperparameters controlling the signal-to-noise ratio, and $\epsilon_{x_t} \sim \mathcal{N}(0,1)$.
BG-De is trained to reverse this process by predicting the noise $\epsilon_\theta(x_t, t)$, enabling the reconstruction of the image.
Unlike the original DDPM, BG-De uses regional masking to focus learning on background regions.
For a polyp detection training sample $\{x, m_{gtb} \}$, where $m_{gtb}$ is a binarized mask indicating the ground truth (GT) bounding boxes of polyps (with $1$ assigned to pixels inside the bounding boxes and $0$ outside), the loss function is modified as:
\begin{equation}
	\mathcal{L}_{\text{BG}} = \mathbb{E}_{x_0,\epsilon,t}  \| (1-m_{gtb}) \odot \left( \epsilon_{x_t} - \epsilon_\theta (x_t, t) \right) \|^2.
\end{equation}

This approach allows BG-De to only model the background (negative) sample distribution, even when polyp samples are present. Notably, the training sets consist solely of polyp images and their corresponding GT bounding boxes, which are standard in detection tasks and \emph{do not require extra data collection efforts}.

\subsection{Detector-guided Adversarial Diffusion Attacker Guiding Ordinary Backgrounds to High-value False Positives}
\label{sec:2.2}
Building on the trained BG-De, we progressively denoise an initialized noise map to synthesize realistic negative endoscopic images $x_0$.
The denoising process is formulated as:
\begin{equation}\small
    x_{t-1} = \mu_\theta ( x_t, t ) + \sqrt{\beta_t}\epsilon,
\label{eq:2}
\end{equation}
where $\mu_\theta(x_t,t)=\frac{1}{\sqrt{\alpha_t}} \left(x_t-\frac{1-\alpha_t}{\sqrt{1-\overline{\alpha}_t}}\epsilon_\theta(x_t,t) \right)$ is the sampled mean value map by BG-De, and $\epsilon \sim \mathcal{N}(0,1)$ is for reparameterization.

However, the generated images are typically ordinary and insufficient to confuse the detector, limiting their training value. To address this, we introduce a perturbation $\eta_t$ at each denoising step to modify the sampling trajectory and guide BG-De toward generating images that exceed the detector's decision boundary. The denoising process becomes:
\begin{equation}
    x_{t-1} = \mu_\theta ( x_t + \eta_t, t ) + \sqrt{\beta_t}\epsilon,
\end{equation}
where $\eta_t$ is a trainable variable initialized as a \emph{zero} matrix.

Inspired by adversarial attack methods~\cite{madry2017towards,ilyas2018black}, we propose the DADA module to optimize the perturbation by targeting the well-trained detector, aiming to induce false positives in a user-defined bounding box $b$. This generates challenging negative samples with misleading interference features. Specifically, we feed $x_{t-1}$ into the detector, treating $b$ as an ``illusory'' ground truth (GT) box, and calculate the detection loss as:
\begin{equation}
    \mathcal{L}_{det} = \mathcal{L}_{cls.}(1,\hat{p}_{\sigma(b)}) + \mathcal{L}_{loc.}(b, \hat{b}_{\sigma(b)}),
\end{equation}
where $\mathcal{L}_{cls.}$ and $\mathcal{L}_{loc.}$ are classification and localization losses, respectively. $\hat{p}_{\sigma(b)}$ and $\hat{b}_{\sigma(b)}$ represent the predicted polyp-class probability and bounding box for the ``illusory'' GT box $b$.
The prediction-GT assignment is related to the detector, \emph{e.g.} Hungarian algorithm~\cite{kuhn1955hungarian} in DETR~\cite{carion2020end}.
The perturbation is then optimized by minimizing the detection loss, and the direction of $\eta_t$ is updated via backpropagated gradients:
\begin{equation}
    \eta_t = \eta_t - \alpha \cdot \text{sgn}( \bigtriangledown_{\eta_t} \mathcal{L}_{det} ),
\label{eq:5}
\end{equation}
where $\text{sgn}(\cdot)$ is the signum function and $\alpha$ is a small step size. By integrating DADA with BG-De, perturbations at each denoising step guide the generation of $x_0$, creating high-value negative samples that significantly confuse the detector.

\subsection{Inpainting Strategy to Maintain Context Consistency}
To enhance the fidelity of the generated image regarding anatomical structures, we use an inpainting strategy that incorporates the real image context to guide local false positive generation. Specifically, we apply ``attack \& inpaint'' within the pre-defined region, while the rest of the regions are kept as the noisy version of the real image.
We define the inpainting region as the same as $b$ described in the Sec.~\ref{sec:2.2}. The final denoising process is formulated as follows:
\begin{equation}
    x_{t-1} = \mu_\theta \left( (1-m_b) \odot x_t^{real} + m_b\odot(x_t + \eta_t), t \right) + \sqrt{\beta_t}\epsilon,
\end{equation}
where $m_b$ is a binarized map, with $1$ assigned to pixels within $b$ and $0$ outside.
This strategy integrates DADA with BG-De, introducing perturbations at each denoising step to generate $x_0$, which maximizes detector confusion while preserving visual integrity.
At last, our method can generate high-value negative samples in a localized region of the real image while maintaining overall context, as shown in Fig.~\ref{fig2}.

\section{Experiments}
\subsection{Datasets and Evaluation Metrics}
We conduct experiments on two datasets: Kvasir~\cite{jha2020kvasir} and an in-house dataset. The Kvasir dataset consists of $1,000$ polyp images, where we generate GT boxes from the provided GT masks. The in-house dataset includes $1,516$ polyp images with GT boxes carefully labeled by two experienced endoscopists from a local hospital. For both datasets, we randomly split them into training, validation, and test sets with an $8:1:1$ ratio. We evaluate detection performance using Precision (P), Recall (R), and F1-score (F1).

\subsection{Implementation Details}
\label{sec:3.2}
For BG-De, we implement the model architecture based on the conventional DDPM~\cite{ho2020denoising} and train it for $320,000$ iterations on two RTX 4090 GPUs with a batch size of $20$. We use YOLO~\cite{jocher2021ultralytics} and DETR~\cite{carion2020end} as detection models, training them according to their official settings. We split the training set into two folds, alternately training BG-De on one while augmenting the other.
For BG-De inference, we set the denoising steps to $1,000$, the final image size to $256 \times 256$, and the perturbation step size $\alpha$ in Eq.~(\ref{eq:5}) to $0.003$.

\subsection{Comparison with State-of-the-arts}
\begin{table}[t]
    \centering
    \caption{Quantitative comparison on Kvasir and our private dataset. The best performance is marked in bold.}
    \label{tab:1}
    \fontsize{8}{9.6}\selectfont  
    \setlength{\tabcolsep}{3pt}  
    \begin{tabular}{clcccccc}
        \toprule
        \multirow{2}{*}{Backbone} & \multirow{2}{*}{Methods} & \multicolumn{3}{c}{Kvasir} & \multicolumn{3}{c}{In-house} \\
        \cmidrule(lr){3-5} \cmidrule(lr){6-8}
        ~ & ~ & P~ & R~ & F1~ & P~ & R~ & F1~ \\
        \hline
        \multirow{6}{*}{YOLO~\cite{jocher2021ultralytics}} 
        & baseline & 0.941 & 0.940 & 0.941 & 0.894 & 0.813 & 0.852 \\
        & APGD~\cite{croce2020reliable} & 0.952 & 0.942 & 0.947 & 0.905 & 0.815 & 0.858 \\
        & FAB~\cite{croce2020minimally} & 0.947 & 0.944 & 0.945 & 0.901 & 0.817 & 0.857 \\
        & Repaint~\cite{lugmayr2022repaint} & 0.951 & 0.946 & 0.948 & 0.911 & 0.823 & 0.865 \\
        & LaMa~\cite{suvorov2022resolution} & 0.953 & 0.947 & 0.950 & 0.915 & 0.819 & 0.864 \\
        & Ours & \textbf{0.983} & \textbf{0.956} & \textbf{0.969} & \textbf{0.942} & \textbf{0.845} & \textbf{0.891} \\
        \hline
        \multirow{6}{*}{DETR~\cite{carion2020end}} 
        & baseline & 0.950 & 0.856 & 0.901 & 0.873 & 0.574 & 0.693 \\
        & APGD~\cite{croce2020reliable} & 0.957 & 0.861 & 0.906 & 0.879 & 0.576 & 0.696 \\
        & FAB~\cite{croce2020minimally} & 0.955 & 0.858 & 0.904 & 0.876 & 0.577 & 0.696 \\
        & Repaint~\cite{lugmayr2022repaint} & 0.962 & 0.862 & 0.909 & 0.885 & 0.582 & 0.702 \\
        & LaMa~\cite{suvorov2022resolution} & 0.963 & 0.859 & 0.908 & 0.886 & 0.585 & 0.705 \\
        & Ours & \textbf{0.980} & \textbf{0.880} & \textbf{0.927} & \textbf{0.918} & \textbf{0.592} & \textbf{0.720} \\
        \bottomrule
    \end{tabular}
\end{table}
We compare our method with four state-of-the-art (SOTA) methods on both Kvasir and the in-house dataset: two adversarial attack methods (APGD~\cite{croce2020reliable}, FAB~\cite{croce2020minimally}) and two inpainting methods (Repaint~\cite{lugmayr2022repaint}, LaMa~\cite{suvorov2022resolution}).
To ensure fairness, we implement all methods as described in Sec.~\ref{sec:3.2}, augment each image once, and train both YOLO and DETR with the original and augmented datasets. Baselines are trained only on the original training set.

\begin{figure}[t]
    \centering
    \includegraphics[width=1\textwidth]{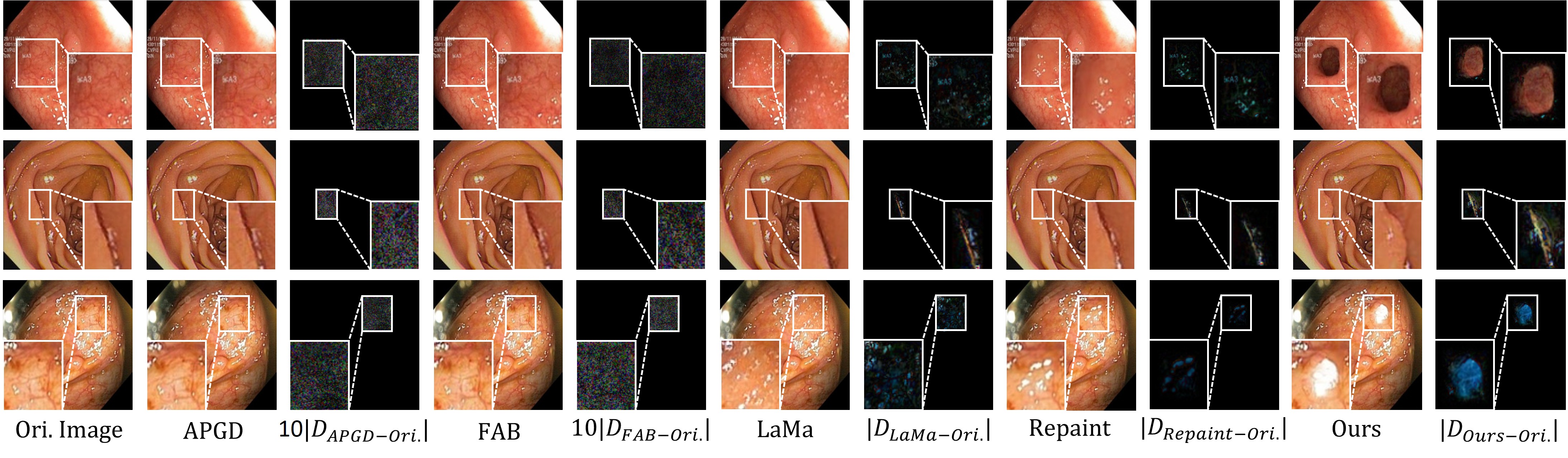}
    \caption{Visualizations of synthesis images by different methods. The white bounding boxes indicate pre-defined attack/inpainting regions. For clarity, we visualize difference maps $D$ between synthesis images and original images.}
\label{fig2}
\end{figure}

As shown in Table~\ref{tab:1}, our method significantly improves the performance of both YOLO and DETR across the two datasets, outperforming all baselines. This improvement mainly stems from our method’s ability to generate high-value negative samples. As shown in Fig.~\ref{fig2}, our method generates challenging negative samples such as circular lumens, colonic folds, and specular highlights, which are likely to mislead the baseline model into producing false positives.
Using these samples as additional training data greatly enhances the performance, especially in terms of precision. In contrast, methods like APGD and FAB generate subtle, noise-like perturbations without clear semantics, while Repaint and LaMa produce ordinary backgrounds that offer limited benefits.

\subsection{Ablation Study}
\begin{table}[t]
    \centering
    \caption{Ablation study of two key components, \emph{i.e.}, BG-De. and DADA.}
    \label{tab:2}
    \fontsize{8}{9.6}\selectfont  
    \setlength{\tabcolsep}{3pt}  
    \begin{tabular}{lcccccccc}
        \toprule
        \multirow{2}{*}{Backbone} & \multicolumn{2}{c}{Components} & \multicolumn{3}{c}{Kvasir} & \multicolumn{3}{c}{In-house} \\
        \cmidrule(lr){2-3} \cmidrule(lr){4-6} \cmidrule(lr){7-9}
        ~ & DADA & BG-De & P~ & R~ & F1~ & P~ & R~ & F1~ \\
        \hline
        \multirow{4}{*}{YOLO} 
        & \XSolidBrush & \XSolidBrush & 0.936 & 0.939 & 0.937 & 0.891 & 0.812 & 0.850 \\
        & \Checkmark  & \XSolidBrush & 0.933 & 0.930 & 0.931 & 0.882 & 0.801 & 0.840 \\
        & \XSolidBrush & \Checkmark  & 0.950 & 0.945 & 0.947 & 0.902 & 0.818 & 0.858 \\
        & \Checkmark  & \Checkmark  & \textbf{0.983} & \textbf{0.956} & \textbf{0.969} & \textbf{0.942} & \textbf{0.845} & \textbf{0.891} \\
        \hline
        \multirow{4}{*}{DETR} 
        & \XSolidBrush & \XSolidBrush & 0.945 & 0.853 & 0.897 & 0.870 & 0.571 & 0.689 \\
        & \Checkmark  & \XSolidBrush & 0.945 & 0.843 & 0.891 & 0.865 & 0.565 & 0.684 \\
        & \XSolidBrush & \Checkmark  & 0.952 & 0.861 & 0.904 & 0.877 & 0.581 & 0.699 \\
        & \Checkmark  & \Checkmark  & \textbf{0.980} & \textbf{0.880} & \textbf{0.927} & \textbf{0.918} & \textbf{0.592} & \textbf{0.720} \\
        \bottomrule
    \end{tabular}
\end{table}
To verify the effectiveness of the two key components, \emph{i.e.}, BG-De and DADA, we train three variants of YOLO and DETR by disabling BG-De and/or DADA.
The comparison results of these methods on Kvasir and our in-house dataset are listed in Table~\ref{tab:2}.
From the first two rows of each detector, we observe that disabling BG-De results in performance degradation compared to their baselines in Table~\ref{tab:1}.
This is primarily because the generator randomly generates both background and polyps, and treating all generated samples as backgrounds introduces label noise and compromises detector training.
As shown in the third row of each detector, introducing BG-De alone provides limited performance improvement, which is mainly because most of the negative samples generated by BG-De are of low-value to the detectors.
Finally, when DADA and BG-De are jointly utilized, both YOLO and DETR achieve the best performance with significant improvements, which demonstrates that DADA and BG-De are highly complementary and must be used in conjunction.

\subsection{Hyperparameter Choices}
To explore the impact of different step sizes $\alpha$ in Eq.~(\ref{eq:5}) on detector performance, we conduct a hyperparameter search using YOLO on both the Kvasir and in-house datasets. We introduce two additional metrics: Fréchet Inception Distance (FID) and False Positive Generation Rate (FPGR). FID measures the distribution discrepancy between generated and real images, while FPGR quantifies the proportion of generated images causing false positives. As shown in Table~\ref{tab:3}, increasing $\alpha$ results in higher FPGR but lower FID, as larger $\alpha$ values introduce more noise into generated images. Based on these results, we set $\alpha = 0.003$, which yields the best detector performance with an F1-score of 0.969 on Kvasir and 0.891 on the in-house dataset.

\begin{table}[t]
    \centering
    \caption{Ablation on perturbation step size $\alpha$. $\uparrow$ and $\downarrow$ denote `higher is better' and `lower is better', respectively.}
    \label{tab:3}
    \fontsize{8}{9.6}\selectfont  
    \setlength{\tabcolsep}{9pt}  
    \begin{tabular}{lcccccc}
        \toprule
        \multirow{2}{*}{$\alpha$} & \multicolumn{3}{c}{Kvasir} & \multicolumn{3}{c}{In-house} \\
        \cmidrule(lr){2-4} \cmidrule(lr){5-7}
        ~ & FID~$\downarrow$~ & FPGR~$\uparrow$~ & F1~$\uparrow$ & FID~$\downarrow$~ & FPGR~$\uparrow$~ & F1~$\uparrow$~ \\
        \hline
        0.001 & \textbf{4.62}  & 0.177 & 0.959 & \textbf{2.78} & 0.216 & 0.868 \\
        0.002 & 5.97 & 0.681  & 0.967 & 3.12 & 0.645 & 0.887 \\
        0.003 & 8.23 & 0.885 & \textbf{0.969} & 3.53 & 0.849  & \textbf{0.891} \\
        0.004 & 9.64 & 0.924 & 0.963 & 3.93 & 0.934 & 0.861 \\
        0.005 & 10.75 & \textbf{0.951}  & 0.954 & 4.20 & \textbf{0.965} & 0.859 \\
        \bottomrule
    \end{tabular}
\end{table}

%
%

\section{Conclusion}
In this work, we introduce a novel approach for generating high-value false positive samples to address the challenge of limited data diversity in polyp detection systems. By integrating a feedback mechanism that targets the detector within the denoising process of a diffusion model, we are able to synthesize realistic, challenging negative samples. This approach, which is the first of its kind in the context of polyp detection, allows for a significant improvement in detector performance by augmenting the training set with these strategically generated false positives. Extensive experiments on both the public Kvasir dataset and our in-house dataset demonstrate that our method enables detectors like YOLO and DETR to achieve state-of-the-art results. Our work contributes a new paradigm for data synthesis in clinical applications, particularly in colorectal cancer screening, where diverse training data is critical for improving detection accuracy and reducing false positives.

\begin{credits}
\subsubsection{\ackname}
This work was supported by the National Natural Science Foundation of China (No: $62401414$).
The in-house dataset used in this work was provided by the research group of Professor Mei Liu from the Department of Gastroenterology at Tongji Medical College of Huazhong University of Science and Technology. We would like to express their sincere gratitude for their valuable contributions and dedicated efforts in collecting and curating the dataset.


\subsubsection{\discintname}
The authors have no competing interests to declare that are relevant to the content of this article.
\end{credits}
%
%
%
%
%
%

	
%
%
%
\bibliographystyle{unsrt}  
\bibliography{references}
\end{document}